\documentclass{article}

\usepackage{arxiv}

\usepackage[utf8]{inputenc} 
\usepackage[T1]{fontenc}    
\usepackage{url}            
\usepackage{booktabs}       
\usepackage{amsfonts}       
\usepackage{nicefrac}       
\usepackage{microtype}      
\usepackage{lipsum}		
\usepackage{graphicx}
\usepackage{natbib}
\usepackage{doi}
\usepackage{multirow}
\usepackage{algorithm}
\usepackage{algorithmic}
\usepackage{amsmath}
\usepackage{tabularx}
\usepackage{listings}
\usepackage{xcolor} 

\title{Zenkai - Framework for Exploring Beyond Backpropagation}

\author{ \href{https://orcid.org/0000-0000-0000-0000}{\includegraphics[scale=0.06]{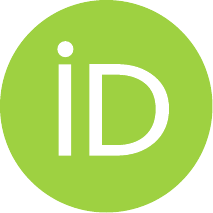}\hspace{1mm}Greg Short}}

\author{
  Greg Short \\
  The Institute of Language and Speech Science \\
  Waseda University \\
  Tokyo, Japan\\
  \texttt{g.short@kurenai.waseda.jp} \\
}

\renewcommand{\shorttitle}{Zenkai - Framework}

\hypersetup{
pdftitle={Zenkai - Framework for Exploring Beyond Backpropagation},
pdfsubject={q-bio.NC, q-bio.QM},
pdfauthor={Greg Short},
pdfkeywords={deep learning, backpropagation, open source framework, backpropagation, target propagation},
}

\begin{document}
\maketitle

\begin{abstract}
	Zenkai is an open-source  framework designed to give researchers more control and flexibility over building and training deep learning machines. It does this by dividing the deep learning machine into layers of semi-autonomous learning machines with their own target and learning algorithm. This is to allow researchers greater exploration such as the use of non-differentiable layers or learning algorithms beyond those based on error backpropagation.
 
    Backpropagation \cite{Rumelhart1986LearningRB} has powered deep learning to become one of the most exciting fields of the 21st century. As a result, a large number of software tools have been developed to support efficient implementation and training of neural networks through the use of backpropagation. While these have been critical to the success of deep learning, building frameworks around backpropagation can make it challenging to implement  solutions that do not adhere to it. Zenkai aims to make it easier to get around these limitations and help researchers more easily explore new frontiers in deep learning that do not strictly adhere to the backpropagation framework. 
\end{abstract}


\section{Introduction}

Deep learning research has boomed over the past 15 years making neural networks into one of the hottest topics in academia and industry. One huge contributor to this proliferation has been open source software tools like Torch7 (\cite{Torch7Collobert2011Torch7AM}), Theano (\cite{Theano2016arXiv160502688short}), Tensorflow (\cite{tensorflow2015-whitepaper}), Chainer (\cite{tokui2019chainer}), and PyTorch (\cite{PyTorchpaszke2019pytorch}). This has made deep learning accessible not only machine learning researchers, but also to researchers in a wide-variety of other fields, to data scientists, to software engineers, and to hobbyists. They are mostly designed to be easy-to-use with an intuitive API that hides the details of automatic differentiation and learning and are also designed to be highly performant. They have also spawned ecosystems of second and third generation libraries and frameworks that have extended the functionality of the core framework like BoTorch (\cite{balandat2020botorch}), extending PyTorch to Bayesian optimization, or Pyro (\cite{bingham2018pyro}), a probabilistic programming framework built with PyTorch, or by improving the quality of life and usability like with Keras (\cite{chollet2015keras}) or Lightning (\cite{Falcon_PyTorch_Lightning_2019}). 

The core tools for deep learning are primarily designed to learn through backpropropagation, the algorithm behind much of the advancements in deep learning \cite{Rumelhart1986LearningRB}. This reliance on backpropagation has been a strength because it has allowed them to highly optimize the performance of the operation and also the usability by abstracting away the details. However, it can also be weakness because it makes it harder to explore different mechanics and learning algorithms for an individual layer, which increases the difficulty of researching topics that do not conform to the backpropagation framework. While in some cases, learning algorithms can still can still be found to cast the problem to the backpropagation framework like with straight-through-estimation \cite{hintonSTE, bengioStochasticNeurons} or like with differentiable decision trees \cite{silva2020optimization}, limits are still exist on what can be learned and how it learns.

Zenkai is an open source tool built on PyTorch \cite{PyTorchpaszke2019pytorch}, which has been developed to overcome these limitations in order to explore other architectures and learning mechanics. This report on Zenkai will be structured as follows. First, the core concepts of Zenkai and the primary design principles of Zenkai will be discussed. Then, the features Zenkai provides will be discussed in more detail. Following that, several experiments will be presented to demonstrate some of the possibilities for what can be done with Zenkai. Lastly, the conclusion and future work will be given. The repository is accessible on Github \footnote{Repository: https://www.github.com/short-greg/zenkai} with documentation \footnote{Documentation: http://zenkai.readthedocs.io} and can be installed with pip \footnote{Installation: \texttt{pip install zenkai}}.

\section{Concept}
\label{sec:concept}


Zenkai is designed to stimulate exploration in deep learning research by dividing a learning machine into layers of semi-autonomous learning machines. It views a learning machine with \texttt{N} hidden layers as being composed of \texttt{N + 1} layers of learning machines depth-wise. And if necessary each of these layers can be potentially be divided into \texttt{M} learning machines width-wise where \texttt{M} is the number of functions generating the outputs for a layer. By dividing it into layers which each have their own \texttt{step()} and \texttt{step\_x()} methods, the researcher can more freely decide on the learning algorithm for each layer or subset of layers.

This concept is inspired by target propagation \cite{autoencodertargetprop2014Bengio, lecunNIPS1988_a0a080f4, autoencodertargetprop1204, 2020targetproptheory} in which targets are propagated backward through the network rather than error gradients and each layer of a neural network has its own target. In target propagation, each target is obtained by using an inverse operation or approximation to the inverse such as the pseudoinverse or a reconstruction operation like in an autoencoder. To illustrate how this works in Zenkai, Code \ref{code:learningmachine} is shown with a dummy class that inherits from the primary class of Zenkai, the LearningMachine.

\label{code:learningmachine}
\begin{lstlisting}[language=Python]
class DummyLearningMachine(LearningMachine):

    def forward(self, x: IO, state: State, release: bool=True) -> IO:
        # use freshen to detach and also set retain_grad to True if not
        # freshened yet
        x.freshen()
        return x

    def assess_y(self, x: IO, t: IO, state: State) -> Assessment:
        return self.loss.assess(x, t, state)

    def accumulate(self, x: IO, t: IO, state: State) -> Assessment:
        # accumulate parameter updates. This step is not necessary

    def step(self, x: IO, t: IO, state: State):
        # update parameters

    def step_x(self, x: IO, t: IO, state: State) -> IO:
        # determine the target for the incoming layer

\end{lstlisting}

The method \texttt{forward()} is the standard forward method inherited from \texttt{nn.Module}. The method \texttt{assess\_y()} is used to evaluate the output of the machine. The method \texttt{step()} updates the parameters of the learning machine. The method \texttt{step\_x()} updates the input which can then be used as the target to the incoming machine. The \texttt{accumulate()} method allows the user to collect changes before being processed similar to \texttt{accGradParameters()}. The learning machines can then be stacked onto one another to form a deep learning machine.

Example Code \ref{code:somewhatdeeplearner} is meant to illustrates that. In this case, it is visible that \texttt{step()} method also serves as the way to propagate to lower layers. Each layer is updated and then the target for the preceding layer is computed with \texttt{step\_x()}. Depending on the layer type, these can be reversed. The \texttt{accumulate()} method is not implemented here and not required. If it is implemented, however, the propagation down through layers will usually need to occur in that method, which removes the ability to alternate the order of \texttt{step\_x()} and \texttt{step()}


\label{code:somewhatdeeplearner}
\begin{lstlisting}[language=Python]
def KindofDeepLearner(LearningMachine):
    ...
    # go backward through the network and update the parameters
    def step(self, x: IO, t: IO, state: State):
        my_state = state.mine(self, x)
        # Here step is done before step_x. That means the parameter update is executed first
        self.layer3.step(my_state.layer2, t)
        # It may be desirable to do step_x first, 
        # like with backpropagation
        t = self.layer3.step_x(my_state.layer2, t)
        # Zenkai also offers convenience methods to reduce the 
        # lines of code like backward()
        self.layer2.step(my_state.layer1, t)
        t = my_state.t1 = self.layer2.step_x(my_state.layer1, t)
        self.layer1.step(x, t)

    # step() must be executed prior to step_x()
    @step_dep('t1')
    def step_x(self, x: IO, t: IO, state: State) -> IO:

        # a dependency is set on step() and it is set so that step() will not be
        # executed automatically. Thus an error will be raised if it hasn't
        return self.layer1.step_x(x, state[self, x, 't1'])
\end{lstlisting}

Because each layer has its own target, pitfalls of using gradient descent such as not being able to train non-differentiable functions can be potentially avoided. Also, as can be seen this makes it easy to have the model parameters updated prior to obtaining the targets for the preceding layer. As mentioned above, target propagation can be used for this when using a trainable layer for reconstruction. The vanilla form of target propagation uses a shallow neural network to predict the targets of the prior layer by approximating the inverse.

\begin{equation}
t_{h} = N^{-1}({t_{h+1}})
\end{equation}

However, gradient descent can also be easily cast to adhere to this framework as the new "target" of layer $h_{i-1}$ can be calculated from the gradient that is backpropagated to the output of layer i. In fact, this approach is typically used to determine the targets for the output layer for target propagation. As can easily be shown, layer $h_{i-1}$ backpropagates the sum of squared errors loss with new target value calculated by updating the output with the gradient.

\begin{equation}
\label{equation:grad}
t_{h} = y_{h} - \nabla N_{h+1}(y_{h})
\end{equation}

And since this can easily be converted to sum of squared errors loss through integration, standard error backpropagation is relatively easy to implement in Zenkai. 

\begin{equation}
\text{SSE} = 0.5 \sum_{i=1}^{n} (y_{h_i} - t_{h_i})^2
\end{equation}

Using this approach would be inferior to just using PyTorch for a regular neural network as it would require more code and lose PyTorch's abstractions. However, this can be useful if some layers make use of gradient descent and others do not or the researcher wants to explore other local loss functions besides SSE for an intermediate layer. Another possibility in Zenkai is to loop over the minibatch and have multiple "local updates" for each "global update".

Zenkai offers more methods for obtaining the target for a layer. For instance, if the layer is non-differentiable, hill-climbing can be used. One possibility for hill climbing is to perturb the inputs and return the input that minimizes the error. In this paper, a variation of that idea is tested. A stochastic layer (using dropout) that precedes the layer is used to generate candidate targets, then the candidate that minimizes the error is chosen as the new target. It is also possible to connect the global loss more directly to the intermediate layer or simply to perturb the outputs of a layer and set the target to the candidate that produces the best results.  While not all possible approaches are guaranteed to lead to minimizing the global cost function, Zenkai gives the researcher more freedom to explore different approaches.

Another benefit of this system is it is also relatively easy to calculate error reduction of the global cost function or a local loss function after each intermediate update of the parameters or the inputs. Global error reduction $(GER)$ can be defined as the reduction in error of the global cost function after an update, 

\begin{equation}
\text{GER} = L_g(N_{h_{pre}-o}(x_{h_{pre}}), t_g) - L_g(N_{h_{post}-o}(x_{h_{post}}), t_g)
\end{equation}

where $ L_g $ represents the global cost, $ N_{h_{pre}-o} $ represents the layers h to o of the network before the update, $ x_{h_{pre}} $ represents the input of \texttt{layer h}, and $ post $ represents them after the update.

Local error reduction can be calculated as the reduction in error for a local cost function (i.e. the loss for an intermediate layer). 
\begin{equation}
\text{LER} =  L_h(N_{h_{pre}}(x_{h_{pre}}), t_h) - L_h(N_{h_{post}}(x_{h_{post}}), t_h)
\end{equation}

where $ L_h $ represents the local loss, $ N_{h_{pre}} $ represents \texttt{layer h} before the update, $ x_{h_{pre}} $ represents the input of \texttt{layer h}, and $ post $ represents them after the update.

This can be helpful to identify the causes of unsuccessful learning. Error reductions tending to be negative might indicate instability which can lead to poor training. Local error reductions being quite small may indicate that the target differences are too small or the learning is too slow.

Through these features, the researcher can experiment with using other techniques such as using decision trees in place of neurons. The researcher can choose to propagate directly from the output layer as in Direct Feedback Alignment. The researcher can also use population-based optimizers to update the layer parameters as has been used in neural network research such as \cite{salimans2017evolution}. There are a multitude of possibilities.

\section{Design and Features}
\label{sec:design}

\subsection{Design Principles}

As exploring new machine architectures and learning algorithms is the primary objective of Zenkai, the primary design objectives for Zenkai are flexibility and ease-of-use. Flexibility is important to be able to test a wide variety of solutions. Ease-of-use is important to keep the difficulty of exploration down. However, increasing flexibility tends to increase the difficulty of usage, because to add flexibility it is often necessary to increase the level of abstractions, but to increase ease-of-use, one must often reduce the level of abstraction. Zenkai mostly follows modular design principles to achieve this aiming for high cohesion and loose coupling. As this requires a balance, it allows the researcher to decide whether to emphasize coupling or cohesion. Also, in order to increase ease-of-use, the systems are implemented to aim for familiarity and consistency. For instance, many of the classes inherit from PyTorch's classes. Also, the learning flow from Torch7 was also mimicked typically with a four step process, \texttt{forward()}, \texttt{accumulate()}, \texttt{step\_x()}, \texttt{step()}. \texttt{accumulate()} can be removed however, and if removed, the order of \texttt{step\_x()} and \texttt{step()} can be reversed.

\subsection{Features}
Zenkai is broken down into six subpackages presently: kaku for the core features and interfaces, kikai for implementations of different types of learning Machines, tansaku for population-based metaheuristics, Mod for utilities that inherit from \texttt{nn.Module} and Utils for functions used by the framework.

\subsubsection{Kaku (i.e. Core)}
\label{subsubsec:core}
Kaku contains the features for defining a LearningMachine (\ref{code:learningmachine}) and all of the components that go into one for updating the parameters, controlling the inputs and outputs, obtaining the targets, assessment, and so on. LearningMachine is the most important class in Zenkai as it defines an interface for implementing learning machines that can be connected to one another.

The core subpackage is imported by simply importing Zenkai '\texttt{import zenkai}'. In addition, there are functionalities in kaku for callbacks for the learning machine, and definitions of criterions that can be used to calculate the loss or value.

\textbf{Components}
\begin{itemize}
    \item \textbf{IO:} Wraps the tensors used for inputs, outputs and targets with a tuple.
    \item \textbf{StepTheta:} Used to update the parameters of the machine
    \item \textbf{StepX:} Used to update the x values to determine the targets of the preceding layer.
    \item \textbf{Assessment:} Contains an evaluation plus whether a cost or utility.
    \item \textbf{State:} Stores the state for the current learning step.
    \item \textbf{LearningMachine:} Learner/nn.Module that inherits from StepTheta and StepX.
    \item \textbf{Individual:} A class for defining one element of a population. Primarily used for metaheuristics.
    \item \textbf{Population:} A class for defining a group of individuals. Primarily used for metaheuristics.
\end{itemize}

\subsubsection{Kikai (Machines)}
\label{subsubsec:kikai}

Kikai contains the class definitions for a variety of LearningMachines, StepThetas or StepXs meant to be the building blocks of larger machines. There are classes for wrapping scikit-learn estimators, for Feedback and Direct Feedback Alignment \cite{FALillicrap2016RandomSF, DFAnøkland2016direct, weightSymmetryImportance, moskovitz2019feedback, refinetti2021align}, for ensembles, for reversible functions, for target propagation, and so on. It also contains classes like GraphLearner which is there to makes it easier to implement compositions of learners.

\subsubsection{Tansaku (Search)}
\label{subsubsec:tansaku}

Tansaku is a collection of modules that allow one to build and execute population-based optimizers using PyTorch primarily to provide more techniques for updating the parameters or obtaining the targets for a LearningMachine. With it, the researcher can flexibly implement different types of such as hill climbing algorithms, genetic algorithms, particle swarm optimization algorithms, and so on. The core classes for it are Individual and Population as well as classes for to modifying an individual or population.

In general, a series of operations like the one in Code \ref{code:optimization} will be used to implement the optimization algorithm. Here you can see a population is generated, it is modified and assessed and finally reduced to an individual.

\label{code:optimization}
\begin{lstlisting}[language=Python]
def climb_hill(self, individual: Individual) -> Individual:
    """Do randomized perturbations on the individual to 
    get a population then get the best individual in the population """
    
    # create a population from the individual
    population = individual.populate(k=8)
    population = self.perturb(population) 
    # evaluate each individual in the population
    population = self.assessor(population) 
    # reduce the population to get the hopefully improved individual
    return self.reducer(population) 
\end{lstlisting}

\subsubsection{Utils and Mod}
\label{subsubsec:utils}

Zenkai also provides a variety of utilities and modules that are used by the rest of Zenkai. These include reversible modules for use by kikai's ReversibleMachine, functions for the setting and getting of model parameters and gradients, and for modules that allow for creating ensemble models, among more.

\section{Example Experiments}
\label{sec:experiments}

In this section, I demonstrate several cases where I've used Zenkai to implement a variety of networks, focusing on ones that make use of techniques other than or in addition to backpropagation. These demonstrations are meant to give a preview of some of the possibilities for learning machines to implement with Zenkai. I have not used any systematic approaches to optimize the hyperparameters so the results in most cases probably do not approximate an upperbound on what is achievable. In some cases, like Feedback Alignment, better results have been demonstrated. Code is given for some of them to demonstrate how the method is implemented with Zenkai, but not all as including code for all of them would lengthen the document considerably. The details of the algorithms proposed in other research are not discussed as that is beyond the scope of this article. 

The results are summarized in Table \ref{table:accuracy}. For each experiment, the network is described, followed by possible use cases and the results. Finally, the network definition is given along with any other necessary details such as algorithms. The MNIST dataset \cite{deng2012mnist} is used for all tests for simplicity and for comparison, even though it may not be ideal in some cases.

\subsection{Baseline}
\label{sec:baseline}

First, a neural network using backpropagation created with the Zenkai framework is tested with two fully-connected "neural" layers. Since Zenkai requires targets to be passed backward, the targets are calculated by the formula in Equation \ref{equation:grad} and the sum of squared error loss is computed and multiplied by 0.5 to backpropagate the gradients to the preceding layer. Details of the architecture and training are given in \ref{subsec:baseline_training_and_architecture}.

\subsection{Decision-Linear}
\label{subsec:decisionlinear}

Decision-Linear demonstrates the potential usage of decision trees as using decision trees could be beneficial because of their interpretability. This shows the use of a machine in which the first layer is composed of an operation other than matrix multiplication. The second layer in the network is a standard linear layer. Details of the architecture and training are given in \ref{subsec:decision_linear_architecture}.

Scikit-learn \cite{scikit-learn} was used for the decision trees. Because \texttt{partial\_fit()} is not available for scikit-learn's decision trees \texttt{fit()} is used instead. However, because in minibatch learning this will overfit to the minibatch, an ensemble of trees is used. Overall, the learning is quite slow since the components were not designed for online updating or optimally designed for outputting multiple values. An alternative is to use one decision tree that outputs multiple values but that has its own downsides like highly correlated outputs. Comparable results to the baseline are achieved, however.

The network uses an ensemble of estimators for the decision tree regressor layer. To form the ensemble for that layer, the nine most recent estimators are used. When fitting for a time step, the oldest estimator is dropped if the number of estimators has reached nine and a new layer is added. The new estimator is then fit. The minibatch size was increased as training progressed until the minibatch size reached the size of the batch.

\begin{algorithm}
\caption{Temporally Dependent Ensemble Updating}
\begin{algorithmic}[1] 
    \IF{ensemble.full()}
        \STATE ensemble.remove\_index(0)
    \ENDIF
    \STATE new\_estimator = copy(base\_estimator)
    \STATE new\_estimator.fit(x, t)
    \STATE ensemble.append(new\_estimator)
\end{algorithmic}
\end{algorithm}

In addition, gradient descent is repeated 40 times on the inputs of the output layer to get the targets for the first layer. Without this, the updates tended to be quite small which resulted in excessively slow training. While this implementation is slow, it is still quite preliminary using components that are not well-designed for this application.

\subsection{Target Propagation: Regularized Least Squares}
\label{subsec:leastsquaresgrad}

Here a neural network is trained that uses target propagation with regularized least squares to determine the target of the previous layer similar to \cite{lecunNIPS1988_a0a080f4} or  \cite{roulet2021target}). A potential benefit of using Least Squares is that it results in larger updates than gradient descent. In this example, it is only used for propagating the target, but it can be used for each layer in the network as well. Details are given in \ref{subsec:target_prop_least_squares_definition}.

To get the target of linear layer, ridge regression is used to find the change in x that minimizes the squared error with Scipy's linear algebra solver method \cite{Scipy2020SciPy-NMeth}. While that is used for the linear operation, gradient descent is used to calculate the targets for activations.

The results do not come close to the baseline, but a lot can be done to improve upon the results such as reducing the "learning rate" over time for target propagation.  Also, it is possible to use least squares for parameter updates and Zenkai provides tools to filter the parameters for that. 

\subsection{Target Propagation: Reconstruction Layer}
\label{subsec:lineartargetprop}
Another approach to target propagation is to make each layer be an autoencoder and to learn an approximation to the inverse (i.e. a reverse operation). The approximation is then used to calculate the target of the incoming layer. This is a more flexible formulation as it can potentially be used to train non-differentiable operations. In this network, I use error backpropagation for the output layer and target propagation for the remaining layers. Also, training is alternated between training the reverse model and training the forward model. Details are given in Section \ref{subsec:semi_target_prop_definition}. The implementation is loosely based on \cite{autoencodertargetprop2014Bengio}.

In \ref{code:targetprop}, the basic code for implementing a layer using target propagation is shown, though this code removes some of the intricacies like training only the forward or reverse layer for simplification. The results were not impressive, however. The learning was a little unstable and the mean-absolute deviation between the inverted y and t for a target propagation layer tend not to shrink. With more sophisticated approaches such as \cite{lee2015difference}, better results should be achievable. Also, learning is alternated between training the reverse model for one epoch and the forward model for the following epoch.

\subsection{Feedback Alignment}
\label{subsec:feedbackalignment}

Feedback alignment \cite{FALillicrap2016RandomSF} is a biologically plausible training algorithm for neural networks that makes use of a randomly generated weight matrix used for backpropagation, which stays fixed. This example shows Zenkai's flexibility in being able to be used when backpropagation does not follow the standard form. Details of the architecture and training are given in \ref{subsec:feedback_alignment_architecture}.

Zenkai makes it more straightforward to implement because the \texttt{step\_x()} method is user-defined. While there are ways to implement this with only PyTorch, it will possibly result in implementing an AutoGrad functor and defining the backward() method to use a random matrix which is passed in.

Feedback alignment updates the weights with the following formula
\begin{equation}
\delta W_i = -((\textbf{B}_{i+1} \delta z_{i+1})*\phi'(z_i))y^{T}_{i-1}
\end{equation}
where $B$ is a randomly generated, constant, $\delta z_{i+1}$ is the gradient from the outgoing layer, $\phi'(z_i))$ is the activation and $y^{T}_{i-1}$ are the outputs of the preceding layer.

These results obtained are not on par when compared to the baseline or in terms of other works but this can be expected with little hyperparameter tuning. Also, Zenkai offers functionality to use other operations besides Torch's \texttt{nn.Linear} in Feedback Alignment.

\subsection{Direct Feedback Alignment (DFA)}
\label{subsec:directfeedbackalignment}

Direct feedback alignment \cite{DFAnøkland2016direct} is a variation of feedback alignment that has a direct connection from each layer to the output layer on the backward pass. This experiment using it illustrates another use of learning that can be challening to implement with standard frameworks due to the non-standard backward pass. Details of the architecture and training are given in \ref{subsec:feedback_alignment_architecture}.

For updates, direct feedback makes use of the matrix \textbf{B} having the shape [N, O] where N is the number of outputs of the layer and O is the number outputs in the network. Otherwise the formula for updating is similar to Feedback Alignment. The implementation is straightforward with Zenkai because the \texttt{step()} method for the network simply calls the step method for each layer without using \texttt{step\_x()}. As with regular feedback alignment, the type of underlying operation can be decided on by the researcher. In this experiment, though, a standard Linear layer was used.

\subsection{Neural Decision}
\label{subsec:neural_decision}

This network is composed of a convolutional network followed by a non-differentiable decision tree classifier for the outermost layer. This demonstrates the possibilities for using a non-differentiable layer in a position that requires propagation to lower layers. Details of the architecture and training are given in \ref{subsec:convolutional_decision_architecture}.

To get around the lack of differentiability, the targets for the convolutional network are computed with a hill climbing algorithm. Candidate targets are generated by the convolutional network by passing each sample through it K times. Since the first layer uses dropout, it is stochastic so K different samples are generated for each instance in the minibatch. The samples that produced the highest average classification rate in the ensemble for each instance for the decision tree layer are chosen to be the targets. The basic algorithm is defined below, though the implementation makes use of vectorized operations rather than loops.

\begin{algorithm}
\caption{Hill-climbing with Stochastic Incoming Layer}
\begin{algorithmic}[1] 
    \REPEAT
    \STATE Repeat the input i K times
    \STATE Compute the output of the incoming layer
    \STATE Reshape the output so that the population dimension is first and the sample dimension is second
    \STATE Choose the output that maximizes the average classification accuracy as the target
    \UNTIL{until looped over all inputs} 
\end{algorithmic}
\end{algorithm}

This produces okay results, but the results did not reach the baseline despite using convolutional operations. Like with target propagation, this resulted in large mean absolute deviations between the outputs and the targets. To improve the results, it is possibly necessary to reduce the variance of the candidate targets that are produced by gradually reducing the dropout probability or some other means, but I have decided to leave that to future work.
\newcolumntype{Y}{>{\raggedleft\arraybackslash}X}

\begin{table}[h]
\label{table:accuracy}
\centering
\caption{Accuracy Results of Experiments}
\begin{tabularx}{\textwidth}{l*{7}{X}} 
\toprule
 & \textbf{Exp 1} & \textbf{Exp 2} & \textbf{Exp 3} & \textbf{Exp 4} & \textbf{Exp 5} & \textbf{Exp 6} & \textbf{Exp 7} \\
\midrule
\textbf{Experiment} & Baseline & Decision-Linear & Least Squares & Linear Target Prop & Feedback Alignment & Direct Feedback Alignment & Neural Decision  \\
\textbf{Accuracy} & 0.969 & 0.967 & 0.941 & 0.929 & 0.958 & 0.940 & 0.939 \\
\bottomrule
\end{tabularx}
\end{table}

\section{Conclusion}
\label{sec:conclusion}

In this document, I have introduced Zenkai, a framework to stimulate deep learning research beyond simple backpropagation. It does this by making it easier and giving more freedom for the researcher to define when and how the hidden layers will be updated and also what the internal mechanics of the layer will be.. This framework aims to provide researchers with more flexibility in architecture and training design for machines with hidden layers. 

I have described the concepts and design principles, which are to provide flexibility and ease-of-use. I have also described the features contained in Zenkai and its subpackages, kikai, tansaku. I then presented several illustrative examples through experiments meant to provide some ideas for what can be accomplished with it with some using non-differentiable operations. These examples showed a variety of possible architectures that can be difficult to implement with standard deep learning frameworks. And there are many more possibilities for what can be implemented with it.

For future work, I plan to create more demonstrations to show how Tansaku can be used effectively. I also intend to expand on the capabilities of the subpackage of \texttt{zenkai.tansaku} and continue to add more features to increase ease of use. I also intend to use Zenkai for work on Deep Neurofuzzy systems, a type of deep learning machine that is difficult to optimize with backpropagation when based on max and min operations in place of addition and multiplication.

\bibliographystyle{unsrtnat}
\bibliography{references}  

\appendix
\section{Learning Machine Definitions}

Below, the definitions for the machines used in the experiments are given.

\subsection{Baseline}

This architecture defines a standard fully connected neural network making use of Zenkai. This approach can be used to connect a neural network that uses gradient descent to some other type of machine.

\label{subsec:baseline_training_and_architecture}
\textbf{Layer 1:} Fully Connected Neural Layer
\begin{enumerate}
    \item Architecture: 128 units, ReLU Activation, BatchNorm
    \item Update: Gradient Descent, Adam Optim, 1e-3 LR, SSE Loss
    \item X Update: N/A
\end{enumerate}
\textbf{Layer 2:} Fully Connected Neural Layer
\begin{enumerate}
    \item Architecture: 32 Units, ReLU Activation, BatchNorm
    \item Update: Gradient Descent, Adam Optim, 1e-3 LR, SSE Loss
    \item X Update: Gradient Descent (from Update)
\end{enumerate}
\textbf{Layer 3:} Fully Connected Neural Layer
\begin{enumerate}
    \item Architecture: 10 Units
    \item Update: Gradient Descent, Optim=Adam, LR=1e-3, Cross Entropy Loss
    \item X Update: Gradient Descent (from Update)
\end{enumerate}

\subsection{Decision-Linear Definition}
\label{subsec:decision_linear_architecture}

This machine has two layers, a layer composed only of decision tree regressors and a standard linear layer. The decesion tree regressors were trained using scikit-learn's \texttt{DecisionTreeRegressor} along with its \texttt{MultiOutput} so for each iteration of training 32 decision trees are learned.

\textbf{Layer 1:} Decision Tree Regressor Layer
\begin{enumerate}
    \item Architecture: Ensemble of 9 / 32 units Decision Trees  -> Voting 
    \item Update: 11 Max Depth, scikit-learn defaults
    \item X Update: N/A
\end{enumerate}
\textbf{Layer 2:} Fully Connected neural Layer
\begin{enumerate}
    \item Architecture: BatchNorm -> 32 Unit Linear -> ReLU Activation, 
    \item Update: Gradient Descent - Adam Optim, 1e-3 LR, Cross Entropy Loss, 128 batch size
    \item X Update: Gradient Descent - 40 iterations
\end{enumerate}

\subsection{Target Propagation: Regularized Least Squares Definition}
\label{subsec:target_prop_least_squares_definition}

The first version of target propagation uses 4 linear layers. The target is propagated backward through the linear layer using the pseudoinverse. For the activations, however, gradient descent was used.

\textbf{Layer 1:} Fully Connected Neural Layer
\begin{enumerate}
    \item Architecture: 128 units Linear -> BatchNorm -> ReLU activation
    \item Update: Gradient Descent - Adam Optim, 1e-3 LR, SSE Loss
    \item X Update: N/A
\end{enumerate}
\textbf{Layer 2:} Fully Connected Neural Layer
\begin{enumerate}
    \item Architecture: 128 units Linear -> BatchNorm -> ReLU activation
    \item Update: Gradient Descent - Adam Optim, 1e-3 LR, SSE Loss
    \item X Update: Regularized Least Squares
\end{enumerate}
\textbf{Layer 3:} Fully Connected Neural Layer
\begin{enumerate}
    \item Architecture: 128 units Linear -> BatchNorm -> ReLU activation
    \item Update: Gradient Descent - Adam Optim, 1e-3 LR, SSE Loss
    \item X Update: Regularized Least Squares 
\end{enumerate}
\textbf{Layer 4:} Fully Connected Neural Layer
\begin{enumerate}
    \item Architecture: 10 units Linear
    \item Update: Gradient Descent - Adam Optim, 1e-3 LR, Cross Entropy Loss
    \item X Update: Regularized Least Squares
\end{enumerate}

\subsection{Semi-Target Prop}
\label{subsec:semi_target_prop_definition}

This variation of target propagation uses autoencoders for each target propagation layer. This machine has 4 linear layers, the outermost of which uses gradient descent for obtaining the incoming target.

\textbf{Layer 1:} Fully Connected Linear Layer 
\begin{enumerate}
    \item Architecture: 128 Units Linear -> ReLU
    \item Update: Gradient Descent, Adam, 1e-3 lr
    \item X Update: N/A
\end{enumerate}
\textbf{Layer 2:} Fully Connected Linear Layer
\begin{enumerate}
    \item Architecture: 128 Units Linear -> ReLU
    \item Update: Gradient Descent, Adam, 1e-3 lr, SSE Loss
    \item Reverse Architecture: 128 Units, BatchNorm, ReLU / 128 Units, BatchNorm, ReLU
    \item Reverse Update: Adam, 1e-3 lr, MSE Loss
    \item X Update: Reverse(t)
\end{enumerate}
\textbf{Layer 3:} Fully Connected Linear Layer
\begin{enumerate}
    \item Architecture: 128 Units Linear -> ReLU
    \item Update: Adam, 1e-3 lr, SSE loss
    \item Reverse Architecture: 128 Units, BatchNorm, ReLU / 128 Units, BatchNorm, ReLU
    \item Reverse Update: Adam, 1e-3 lr, MSE Loss
    \item X Update: Reverse(t)
\end{enumerate}
\textbf{Layer 4:} Fully Connected Linear Layer
\begin{enumerate}
    \item Architecture: 10 units Linear
    \item Update: Gradient Descent, Adam, 1e-3 lr
    \item X Update: Gradient descent
\end{enumerate}

Here is an example of the code that can be used to implement a target propagation layerf or this example.

\label{code:targetprop}
\begin{lstlisting}[language=Python]
class LinearTargetPropLearner(zenkai.LearningMachine, TargetPropStepX):

    def __init__(
        self, in_features: int, out_features: int, 
        lr: float=1e-3, reduction: str="mean", loss="cross_entropy",
        name: str="TargetProb"
    ):
        super().__init__()
        self.layer = nn.Sequential(
            nn.Linear(in_features, out_features),
            nn.BatchNorm1d(out_features),
            nn.ReLU(),
        )
        reverse_layer = nn.Sequential(
            nn.Linear(out_features, in_features),
            nn.BatchNorm1d(in_features),
            nn.ReLU(),
            nn.Linear(in_features, in_features),
            nn.BatchNorm1d(in_features),
        )
        self._target_prop_learner = GradLearner(
            reverse_layer, ThLoss('mse', reduction="mean"), OptimFactory('adam', 1e-3), "mean"

        )
        self._step_theta = zenkai.kikai.GradStepTheta(self, zenkai.itadaki.adam(lr=lr), reduction)
        self._loss = zenkai.ThLoss(loss, reduction=reduction, weight=0.5)

    def step_target_prop(self, x: IO, t: IO, y: IO, state: State):
        self._target_prop_learner.step(y, x, state)

    def accumulate(self, x: IO, t: IO, state: State):
        self._step_theta.accumulate(x, t, state)
        x_prime = self._target_prop_learner(state[self, 'y'], state)
        self._target_prop_learner.accumulate(state[self, 'y'], x.detach(), state)
    
    def step_x(self, x: IO, t: IO, state: State) -> IO:
        x_prime = self._target_prop_learner(t, state, release=True)
        return x_prime

    def assess_y(self, y: IO, t: IO, reduction_override: str = None) -> AssessmentDict:
        return self._loss.assess_dict(y, t, reduction_override)
    
    def forward(self, x: IO, state: State, release: bool = True) -> IO:
        x.freshen()
        y = state[self, 'y'] = self.layers(x)
        return y.out(release)

    def step(self, x: IO, t: IO, state: State):
        
        self._step_theta.step(x, t, state)
        self.step_target_prop(x, t, state[self, 'y'], state)


\end{lstlisting}

\subsection{Feedback Alignment}
\label{subsec:feedback_alignment_architecture}

Feedback Alignment makes use of a randomly generated weight matrix used in the backward pass that remains constant. This is a 3 layer fully-connected network implementation of feedback alignment.

\textbf{Layer 1:} Fully Connected FA Layer 
\begin{enumerate}
    \item Architecture: 32 output units, ReLU activation
    \item Update: Gradient descent
    \item X Update: N/A
\end{enumerate}
\textbf{Layer 2:} Fully Connected FA Layer
\begin{enumerate}
    \item Architecture: 10 output units
    \item Update: Gradient Descent
    \item X Update: Gradient descent using B
\end{enumerate}

\subsection{Direct Feedback Alignment}
\label{subsec:direct_feedback_alignment_architecture}

Direct Feedback Alignment makes use of a randomly generated weight matrix used in the backward pass that remains constant. In addition to that, each layer is directly connected to the output on the target pass. This is a 3 layer fully-connected network implementation of Direct Feedback Alignment.

\textbf{Layer 1:} Fully Connected DFA Layer 
\begin{enumerate}
    \item Architecture: 32 units Linear -> BatchNorm -> ReLU
    \item Update: Cross Entropy Loss, 1e-3 LR
    \item X Update: N/A
\end{enumerate}
\textbf{Layer 2:} Fully Connected DFA Layer
\begin{enumerate}
    \item Architecture: 32 units Linear -> BatchNorm -> ReLU
    \item Update: Cross Entropy Loss, 1e-3 LR
    \item X Update: N/A
\end{enumerate}
\textbf{Layer 3:} Fully Connected FA Layer
\begin{enumerate}
    \item Architecture: 10 units Linear
    \item Update: Cross Entropy Loss, 1e-3 LR
    \item X Update: N/A
\end{enumerate}

Here is an example of code used for implementing Direct Feedback Alignment.

\label{code:dfalearner}
\begin{lstlisting}[language=Python]
def DFALearner(LearningMachine):
    ...

    def __init__(self, in_features, hidden_features1, hidden_features2, out_features):

        super().__init__()
        self.layer3 = FALearner(nn.Linear(hidden_features2, out_features))
        self.layer2 = DFALearner(nn.Linear(hidden_features1, hidden_features2))
        self.layer1 = DFALearner(nn.Linear(in_features, hidden_features2))
        self.loss = ThLoss('cross_entropy')

    def assess_y(self, x, t, state, reduction_override: ) -> As
        return self.loss.assess_dict(x, t, reduction_override)
        
    def step(self, x: IO, t: IO, state: State):
        # go backward through the network. 
        my_state = state.mine(self, x)
        # Here step is done before step_x. That means the update is executed first
        self.layer3.step(my_state.layer2, t)
        self.layer2.step(my_state.layer1, t)
        self.layer1.step(x, t)
        state[self, x, 'stepped'] = True

    # requires that step() is executed first
    @step_dep('stepped')
    def step_x(self, x: IO, t: IO, state: State) -> IO:
        # backpropgates from the first layer
        return self.layer1.step_x(x, state[self, x, t])
\end{lstlisting}

\subsection{Convolutional - Decision}
\label{subsec:convolutional_decision_architecture}

This is a three layer fully-connected network built from a convolutional network with one convolutional layer and a fully-connected layer followed by a decision tree classifier.

\textbf{Layers 1 and 2:} Convolutional neural network
\begin{enumerate}
    \item Architecture Layer 1: (32 units, 4 kernel, 2 stride) -> BatchNorm -> ReLU
    \item Architecture Layer 2: Linear -> BatchNorm
    \item Update: MSE Loss, Adam Optim, Batch Size=128
    \item X Update: N/A
\end{enumerate}
\textbf{Layer 2:} Decision Tree Classifier
\begin{enumerate}
    \item Architecture: 9 ensemble / 10 Max-depth 1 categorical output Decision Tree Classifier
    \item Update: Scikit-learn defaults
    \item X Update: Hill Climbing with Stochastic Incoming Layer
\end{enumerate}







\end{document}